\documentclass[runningheads]{llncs}
\usepackage{graphicx}
\usepackage{amsmath}
\usepackage{amssymb}
\usepackage{booktabs}

\usepackage[thinlines]{easytable}
\usepackage{enumitem}
\usepackage{multirow}
\usepackage{tabu}
\usepackage[dvipsnames]{xcolor}
\usepackage{transparent}

\usepackage{tabulary}
\usepackage{amsfonts}       
\usepackage{nicefrac}       
\usepackage{microtype}      
\usepackage{enumitem}
\usepackage{tabularx}
\usepackage{adjustbox}
\usepackage{subfig}
\usepackage{lipsum}
\usepackage[linesnumbered,ruled,vlined]{algorithm2e}

\newcommand{\figref}[1]{Fig.~\ref{#1}}
\newcommand{\tabref}[1]{Table~\ref{#1}}

\newcommand{\algref}[1]{Alg.~(\ref{#1})}
\newcommand{\secref}[1]{Sec.~\ref{#1}}

\usepackage{tikz}
\usepackage{comment}
\usepackage{amsmath,amssymb} 
\usepackage[accsupp]{axessibility}  

\usepackage[pagebackref=true,breaklinks=true,letterpaper=true,colorlinks,bookmarks=false]{hyperref}
\hypersetup{
     colorlinks   = true,
     citecolor    = green
}
\hypersetup{linkcolor=red}

\begin{document}

\newcommand\blfootnote[1]{%
  \begingroup
  \renewcommand\thefootnote{}\footnote{#1}%
  \addtocounter{footnote}{-1}%
  \endgroup
}

\pagestyle{headings}
\mainmatter
\def\ECCVSubNumber{5114}  

\title{
Tracking by Associating Clips
} 

\titlerunning{Tracking by Associating Clips}
%
\author{First Author\inst{1}\orcidID{0000-1111-2222-3333} \and
Second Author\inst{2,3}\orcidID{1111-2222-3333-4444} \and
Third Author\inst{3}\orcidID{2222--3333-4444-5555}}
\author{
Sanghyun Woo\inst{1} \and
Kwanyong Park\inst{1} \and \\
Seoung Wug Oh\inst{2} \and 
In So Kweon\inst{1} \and
Joon-Young Lee\inst{2}
}
\institute{
KAIST \and Adobe Research}
\authorrunning{S. Woo et al.}
\maketitle

\begin{abstract}
The tracking-by-detection paradigm today has become the dominant method for multi-object tracking and
works by detecting objects in each frame and then performing data association across frames.
However, its sequential \textit{frame-wise matching} property fundamentally suffers from the intermediate interruptions in a video, such as object occlusions, fast camera movements, and abrupt light changes. Moreover, it typically overlooks temporal information beyond the two frames for matching.
In this paper, we investigate an alternative by treating object association as \textit{\textbf{clip-wise matching}}.
Our new perspective views a single long video sequence as multiple short clips, and then the tracking is performed both within and between the clips.
The benefits of this new approach are two folds.
First, our method is robust to tracking error accumulation or propagation, as the video chunking allows bypassing the interrupted frames, and the short clip tracking avoids the conventional error-prone long-term track memory management.
Second, the multiple frame information is aggregated during the clip-wise matching, resulting in a more accurate long-range track association than the current frame-wise matching.
Given the state-of-the-art tracking-by-detection tracker, QDTrack, we showcase how the tracking performance improves with our new tracking formulation.
We evaluate our proposals on two tracking benchmarks, TAO and MOT17 that have complementary characteristics and challenges each other.

\end{abstract}

\section{Introduction}
Discriminating the identity of multiple objects in a scene and providing individual trajectories of their movements over time, namely multi-object tracking,
is one of the fundamental computer vision problems, imperative to tackle many real-world problems, \textit{e.g.} autonomous driving and surveillance.
Despite being a rather classical vision task, it is still challenging to design a robust multi-object tracker capable of tracking a time-varying number of objects moving through unconstrained environments in the presence of many other complexities.

Early studies approached the multi-object tracking problem by breaking it into multiple sub-problems that could be tackled individually, 
typically starting with object detection, followed by association, track management, and post-processing~\cite{andriyenko2011multi,andriyenko2012discrete,bewley2016simple,fortmann1983sonar,rezatofighi2015joint,streit1994maximum}. 
Ever since, this tracking-by-detection paradigm has become the standard approach for multi-object tracking, and most of the state-of-the-art trackers follow this scheme~\cite{bergmann2019tracking,wang2020towards,liang2020rethinking,zhang2021fairmot,pang2021quasi,zeng2021motr,zhang2021bytetrack}.

\begin{figure*}[t]
    \centering 
    \includegraphics[width=1\textwidth]{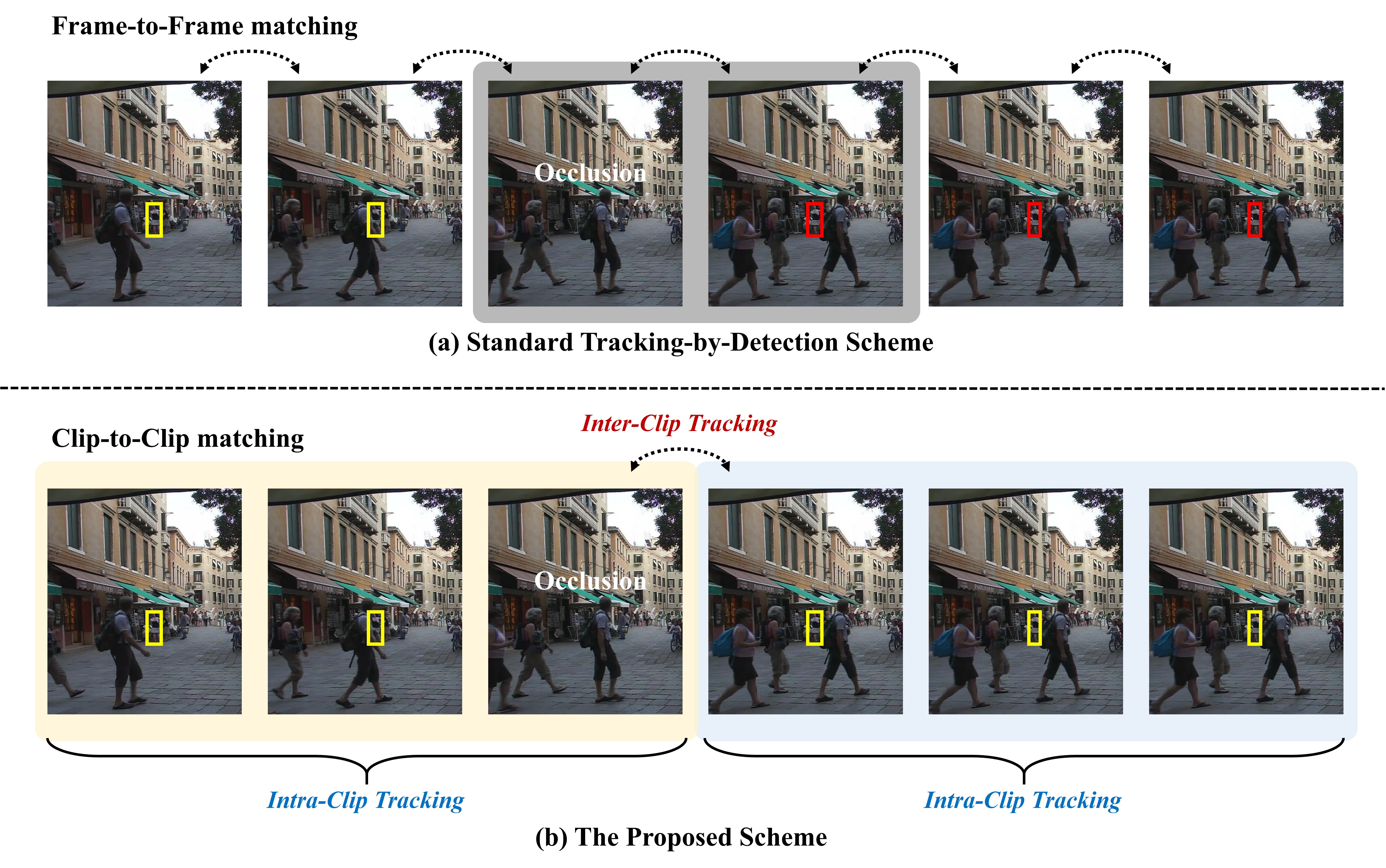}
    \captionsetup{font=footnotesize}
    \caption{
    \textbf{The proposed Clip-to-Clip matching.}
    (a) The current tracking paradigm sequentially matches the object instances frame-by-frame, which is vulnerable to intermediate interruptions in a video and cannot faithfully exploit the temporal information during the matching, resulting in tracker drifting.
    (b) Unlike the standard frame-wise matching scheme, we introduce a new clip-wise tracking method.
    Our approach is robust to random interference in a video, as we can skip those frames by chunking the video into multiple clips.
    Moreover, the multiple frame information is exploited during the clip-wise matching.
    With our proposal, the object can be seamlessly tracked over time.
    }
    \label{fig:teaser}
\end{figure*}

The tracking-by-detection scheme is essentially a sequential frame-by-frame association approach (see~\figref{fig:teaser}-(a)).
In practice, it is achieved by matching the temporally smoothed history, which can be either new locations based on the past motion records~\cite{zhang2021bytetrack} or moving averaged RE-ID features of the trajectories~\cite{pang2021quasi}, with the current predictions.
However, this sequential frame-based matching fundamentally suffers from two prominent limitations.
First, the error accumulation or propagation cannot be handled properly.
For example, sudden object motion pattern changes or fast camera movements significantly disturb the previous motion records.
Also, object deformation such as occlusion or blur corrupts the moving averaged RE-ID features, leading to track drifting.
The phenomenon becomes even more severe when the input video frame rate is low.
Second, it does not consider the overall temporal context during matching.
Looking at only two frames for matching essentially has ambiguity, resulting in track fragmentation.

To effectively utilize the temporal information and make the tracker robust to intermediate interruptions in a video, we reformulate the standard tracking-by-detection scheme as a clip-to-clip matching problem (see~\figref{fig:teaser}-(b)).
In particular, we chunk a video into multiple short clips and perform the tracking at the clip level.
The video chunking allows skipping challenging frames for tracking, and the short-clip tracking avoids long-term track memory management such as long sequence temporal smoothing of RE-ID features along the video.
Furthermore, the association between clips utilizes multi-frame information, which is more robust than the previous frame-based association.
While not being sensitive to any specific design of tracker, we build our proposal upon the state-of-the-art RE-ID-based tracking-by-detection paradigm, QDTrack~\cite{pang2021quasi}, as it can handle both the low~\cite{dave2020tao} and high frame rate~\cite{milan2016mot16} video inputs better than the motion-based trackers~\cite{zhou2020tracking,liang2020rethinking,zhang2021fairmot,wu2021track,zhang2021bytetrack}.

We define two new basic tracking operations to implement the clip-wise tracker: intra- (within-) and inter- (between-) clip tracking.
We target \textit{semi-online scenarios} throughout this paper. Thus we only allow the intra-clip association to perform matching irrespective of the frame order.
Meanwhile, inter-clip tracking incrementally merges the clip predictions until the complete video-level object trajectories are all retrieved.
It requires associating objects at a track level, and thus the temporal information plays an important role here.

We investigate various viable implementations for both intra- and inter- clip tracking.
Specifically, for the intra-clip tracking, we explore two feasible association approaches, directional and direction-free matching~\cite{zhang2017multi}.
For the inter-clip tracking, we consider the IoU-based chaining~\cite{peng2020chained} and temporal averaged feature matching.
We further improve the inter-clip matching by designing a novel transformer-based clip tracker.
It learns to temporally attend to all the past track embeddings of each object and predicts representative embeddings.
These temporally attended embeddings are matched to associate the clip-level predictions.
To make the model more robust at test time, we simulate hard positives and negatives during training: mislocalization and track drifting.
We apply our inter-clip tracker recursively over time and merge the clip predictions into global object tracks incrementally.

With our proposals, we achieve state-of-the-art results on the challenging large vocabulary object tracking benchmark, TAO.
We also validate our approach on the MOT17 benchmark to focus on the tracking performance more thoroughly.
Compared to the baseline, we observe clear improvement in the association performance.
Finally, we conduct extensive ablation studies and confirm that the proposals are effective and generic.

\section{Related Work}

\noindent{\textbf{Multi-Object Tracking.}}
\sloppy The tracking-by-detection paradigm~\cite{ramanan2003finding} dominates the current state-of-the-art multi-object tracking frameworks~\cite{leal2017tracking}. 
It detects objects first and associates them frame-by-frame across time~\cite{andriyenko2011multi,andriyenko2012discrete,bewley2016simple,fortmann1983sonar,rezatofighi2015joint,streit1994maximum}.
Recently, deep learning has contributed significantly to improving the performance of multi-object tracking approaches by focusing on designing better detectors~\cite{ren2015faster,redmon2016you} or developing more effective association objectives~\cite{hu2019joint,leal2016learning,sadeghian2017tracking,wang2020towards,wojke2017simple,pang2021quasi}.
Several works also have shown advances regarding detection and tracking as a joint learning task~\cite{feichtenhofer2017detect,sun2019deep,wu2021track,zhou2020tracking}. Nonetheless, these methods often formulate the multi-object tracking problem only with two consecutive frames and dismiss long-term temporal information, which is crucial for tackling various challenges in tracking.

While sequential frame-based matching is efficient and has shown promising results, they are fundamentally vulnerable to intermediate interruptions in a video and miss rich temporal information during matching.
Unlike these methods, we target a semi-online scenario where we allow slight lagging and utilize several consecutive frames (\textit{i.e.} clip) as a whole for the association.
We show that our proposal is robust to intermediate interruptions in a video and produces better association quality.

\vspace{2mm}
\noindent{\textbf{Clip-level Modeling in Video.}}
The simultaneous process of multiple frames (\textit{i.e.} clip-level modeling) is a promising research direction in many video tasks including video instance segmentation~\cite{athar2020stem,wang2021end,hwang2021video,wu2021seqformer}, video object segmentation~\cite{park2022per}, optical flow estimation~\cite{janai2018unsupervised}, and depth estimation~\cite{watson2021temporal}.
Per-clip models enjoy the opportunities to extract rich temporal information from multiple frames, leading to effectively tackle ambiguities in videos (\textit{e.g.} occlusions or blur).
Most existing methods focus on extracting temporal information to improve recognition or segmentation quality, and there has been little consideration to the association part.  

Here, we present to associate objects at clip-level, performing the tracking both within and in-between the clips.
The clip-level matching provides two clear advantages.
First, by chunking, the matching within the clips essentially becomes short-term tracking, avoiding long-term history averaging or memory management.
Second, the matching in-between the clips allows exploiting the temporal context, resolving the temporally-local ambiguities of matching between the two consecutive frames.
To achieve better long-term association with accurate inter-clip matching, we design a novel transformer architecture.

\vspace{2mm}
\noindent{\textbf{Transformers for Tracking.}}
Recently, transformers~\cite{vaswani2017attention} have shown impressive results in many computer vision tasks, such as image classification~\cite{dosovitskiy2020image,liu2021swin}, object detection~\cite{carion2020end,zhu2020deformable}, segmentation~\cite{zheng2021rethinking}, and image generation~\cite{parmar2018image}.
There are also several methods~\cite{sun2020transtrack,meinhardt2021trackformer,zeng2021motr} adopting the transformers for multi-object tracking by extending DETR frameworks~\cite{carion2020end}.
However, these methods are still limited at utilizing short-term temporal information and rather adopt conventional heuristics such as Intersection over Union (IoU) matching~\cite{sun2020transtrack}, formulate the problem as a two frames task~\cite{meinhardt2021trackformer}, or rely on the frame-by-frame evolving track quries~\cite{zeng2021motr}.

Apart from the previous works, we set up the clip-based tracking scenario.
We see that the key to accurate inter-clip matching is to generate discriminative clip-level representations.
Our main idea is to use a Transformer as a track history summarizer.
In practice, given a short track sequence of object instance (\textit{i.e.} intra-clip tracking results), we input this track sequence to the transformer and get the condensed feature. Finally, this track summary feature is used for the subsequent inter-clip matching.

\begin{algorithm}[t!]
    \KwIn{
        A video sequence $\texttt{V} = (f_{1},\ldots,f_{N})$; 
        object detector \texttt{Det}; 
        clip size $\texttt{C}_{S}$;
        clip interval $\texttt{C}_{I}$;
    }
    \KwOut{
        Tracks $\mathcal{T}$ of the video
    }
    \Begin
    {   
        Initialization: $\mathcal{T} \leftarrow \emptyset$\\
        \For{$i = 1 : C_{I} : N$} 
        {
            \texttt{/* Sample clip */} \\
            $\mathcal{C} \leftarrow \texttt{V}[i:i+\texttt{C}_{S}]$ \\
            
            \texttt{\\}
            
            \texttt{/* Predict detection boxes \& scores */} \\
            $\mathcal{D}_{clip} \leftarrow \emptyset$\\
            \For{frame $f_{k}$ in  C}
            {
                $\mathcal{D}_{k} \leftarrow \texttt{Det}(f_{k})$ \\
                $\mathcal{D}_{clip} \leftarrow \mathcal{D}_{clip} \cup \mathcal{D}_{k}$ \\
            }
            
            \texttt{\\}
            
            \texttt{/* Intra clip association*/} \\
            Initialization: $\mathcal{T}_{intra} \leftarrow \emptyset$ \\
            Associate $\mathcal{T}_{intra}$ and $\mathcal{D}_{clip}$
            
            \texttt{\\}
            \texttt{/* Inter clip association*/} \\
            Associate $\mathcal{T}$ and $\mathcal{T}_{intra}$
        }
    }
    \caption{Tracking by Associating Clips}
    \label{alg:clip_track}
\end{algorithm}

\section{Tracking by Associating Clips}

We propose a simple yet effective data association method, Tracking by Associating Clips.
Unlike the standard tracking-by-detection scheme, which matches object instances frame-by-frame, we instead conduct association at a clip level as shown in~\figref{fig:teaser}.
We provide the pseudo-code in~\algref{alg:clip_track}.

The input is a video sequence $\texttt{V}$ and an object detector $\texttt{Det}$.
Also, we set two hyper-parameters, clip size $\texttt{C}_{S}$ and clip interval $\texttt{C}_{I}$.
The output is the tracks $\mathcal{T}$ of the video, and each track contains the bounding box and identity of the object in each frame.

After sampling the clip from the video, we predict the detection boxes and scores using the detector $\texttt{Det}$ for each frame.
Then, the first association is performed to link the detection boxes in the clip $\mathcal{D}_{clip}$, which results in local clip-level tracks $\mathcal{T}_{intra}$.
The second association is performed between the local clip-level tracks $\mathcal{T}_{intra}$ and the global video-level tracks $\mathcal{T}$.
The global video-level tracks $\mathcal{T}$ keep the history of the tracked objects and use this information to associate with the local clip-level tracks $\mathcal{T}_{intra}$.
The object tracks are incrementally expanded over time by repeating the intra- and inter-clip associations.
Notably, the proposal degenerates to the standard tracking-by-detection scheme when the clip window size reduces to 1 without overlaps in between.

In this paper, we built our proposal upon the recently presented QDTrack~\cite{pang2021quasi}, given its strong Re-ID association capability.
It is more robust than motion-based trackers~\cite{zhou2020tracking,liang2020rethinking,zhang2021fairmot,wu2021track,zhang2021bytetrack} to target both the low-frame-rate (\textit{e.g.} 1FPS, TAO~\cite{dave2020tao}) and high-frame-rate (\textit{e.g.} 30FPS, MOT17~\cite{milan2016mot16}) video inputs. In the following, we explore effective implementations of intra- and inter-clip association methods.

\subsection{Intra Clip Association}
\label{sec:intra}
\begin{figure*}[t]
    \centering 
    \includegraphics[width=1\textwidth]{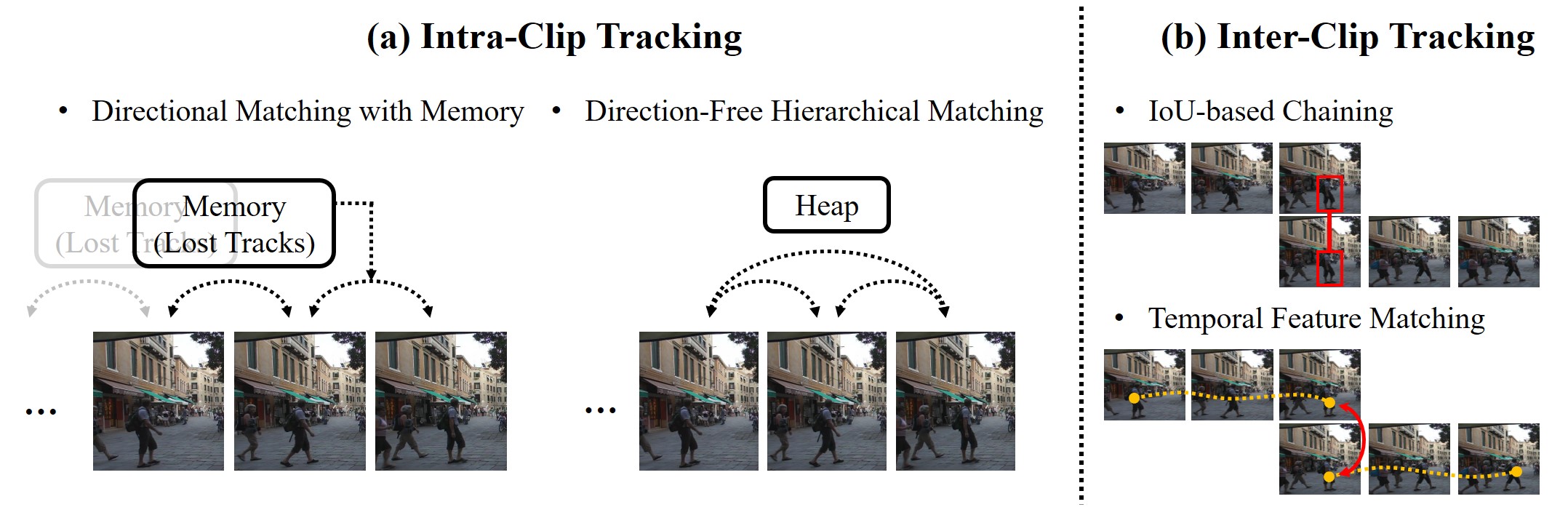}
    \captionsetup{font=footnotesize}
    \caption{
    \textbf{Two Basic Operations for Clip-based Tracking.}
    We define intra- and inter- clip tracking for instantiating the clip-based tracker.
    For intra-clip tracking, we consider both directional and direction-free matching approaches.
    The former is a standard frame-by-frame matching with a memory mechanism.
    The latter is implemented using the heap structure and hierarchically clusters the detection boxes across the frames without considering the frame order~\cite{zhang2017multi}.
    For the inter-clip tracking, we examine IoU-based chaining~\cite{peng2020chained} and temporal average-pooled feature matching.
    }
    \label{fig:intra_inter}
\end{figure*}

Given the detection boxes of each frame in the clip, we examine two possible implementations, directional and direction-free matching (see~\figref{fig:intra_inter}-(a)).

\vspace{2mm}
\noindent{\textbf{Directional Matching.}}
The standard way to associate the clip-level object predictions is to link them sequentially.
We see that the matching direction, either left to the right or right to the left, provides similar performance; thus, we only consider the former case.
To address object occlusions within the clip, we employ track rebirth~\cite{wojke2017simple,chen2018real} using a memory mechanism.

\vspace{2mm}
\noindent{\textbf{Direction-Free Matching.}}
Within the clip, the motion patterns and appearances of objects are quite similar due to the inherent redundancy in a video. 
Given this fact, we can also associate the objects in a direction-free manner.

Specifically, we employ the heap-based hierarchical clustering algorithm~\cite{zhang2017multi}.
At the beginning, we consider that every detection box in all frames has its own cluster. We calculate their pair-wise appearance distances~\cite{pang2021quasi}.
We avoid the association of the objects in the same frame by setting their distance to be infinity.
Subsequently, every cluster distance will be inserted into a priority queue based on a heap data structure.
Afterward, the cluster pair associated with the smallest distance in the priority queue will be popped and will be merged together.
For this new cluster, distances to other clusters have to be computed and inserted into the priority queue.
To set the distance between the track pairs, we search the minimum distance.
The clustering is repeated until there are only distances greater than a threshold left.

For the intra-clip tracking, which is essentially short-term tracking, we observe that the standard directional matching with the lost track management is already competitive. It outperforms the strong heap-based hierarchical clustering algorithm~\cite{zhang2017multi} both in accuracy and efficiency.

\subsection{Inter Clip Association}
\label{sec:inter}
This section explores how to match and merge the current clip-level predictions into the global video-level tracks.
We consider the following two feasible implementations (see~\figref{fig:intra_inter}-(b)).

\vspace{2mm}
\noindent{\textbf{Iou-Based Matching.}}
We adopt IoU-based chaining~\cite{peng2020chained} that associates object tracks based on the IoU in the same frame.
However, this method has three apparent drawbacks.
First, it misses linking all the object tracks not revealed in the overlapping frame, and this phenomenon becomes even more severe as clip window size increases.
Second, as the number of overlapping frames increases, the single-frame-based IoU matching does not guarantee optimal matching.
Third, the method is basically not applicable when there are no overlapping frames.

\vspace{2mm}
\noindent{\textbf{Temporal Feature Matching.}}
Free from the above limitations, an alternative is based on feature matching.
As a simple baseline, we employ temporally average-pooled features for the association.

While simple, temporal feature matching improves over the frame-based tracking baseline and outperforms the IoU-based chaining~\cite{peng2020chained}.
However, we found that it is not always the case; For the low frame rate video inputs (\textit{i.e.} TAO), which inherently include severe appearance change across the frame, the simple averaged temporal matching is inferior to the baseline (see~\tabref{tab:intra_inter_abl}).
To improve the clip-level association quality further, we designed a new Transformer-based clip tracker, described in the following section.

\subsection{Clip Tracker}
\label{sec:clip_tracker}
\begin{figure*}[t]
    \centering 
    \includegraphics[width=1\textwidth]{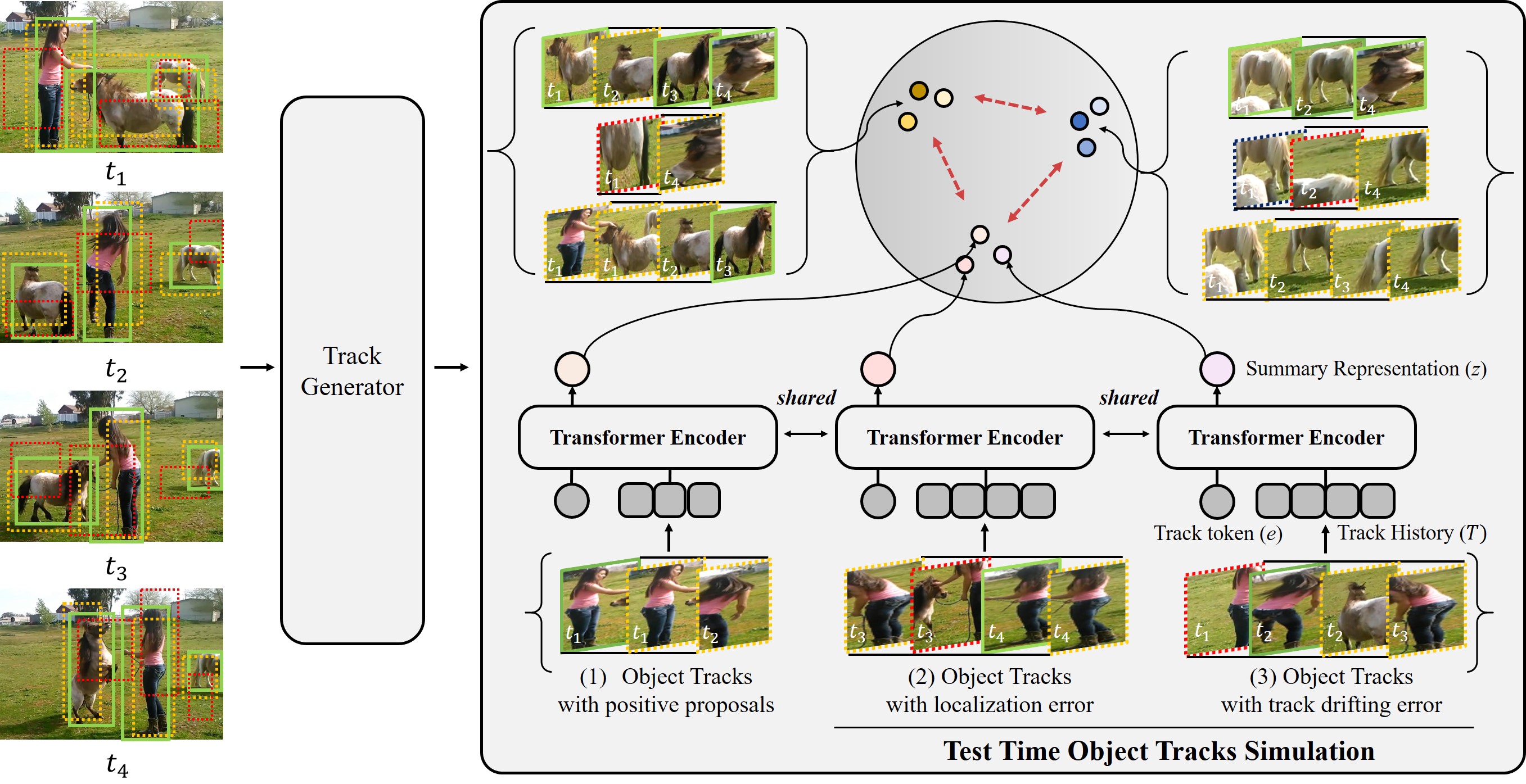}
    \captionsetup{font=footnotesize}
    \caption{
    \textbf{The overview of Clip Tracker.}
    The solid \textcolor{OliveGreen}{\textbf{green}} line indicates the original ground truth boxes of each instance.
    The \textcolor{Dandelion}{\textbf{yellow}} and \textcolor{RubineRed}{\textbf{red}} dotted lines are positive and negative proposals, respectively.
    The transformer learns a \textit{weighted temporal pooling} function of track history in a data-driven manner.
    \textit{Best viewed in color.}
    }
    \label{fig:clip_tracker}
\end{figure*}

The most challenging aspect of the inter-clip association is to generate a discriminative \textit{clip-level} embedding for any object tracks.
Our key idea for this problem is to use transformer as a track history summarizer (see~\figref{fig:clip_tracker}).
By design, the transformer can process and relate set-based elements.
Here, we consider the intra-clip results (\textit{i.e.} short object track sequence) as an input for the transformer, and we attempt to temporally-pool their feature information into a condensed feature.
During training, the clip tracker learns to produce a good summary of the given object track history, which can link the clip-level predictions over time at test time.

In practice, we design the clip tracker using transformer encoder layers.
For object $p$, the clip tracker takes all the feature embeddings in the track $\mathcal{T}^{p} = \{x^{p}_{t_{s}},\dots,x^{p}_{t_{e}}\} \in R^{L\times C}$
and an extra learnable track token $e \in R^{C}$ as an input.
Here, ${t_{s}}$ and ${t_{e}}$ denote the track initiation and termination time, respectively.
$L$ and $C$ are the track length and the dimension of the feature embeddings, respectively.
This input sequence, $[e, \mathcal{T}^{p}] \in R^{(L+1)\times C}$, is forwarded to the transformer, and we take the output of the track token, $z^{p} \in R^{C}$, as a condensed representation of the given object's track history.

This summary representation should be representative enough to match the same object instance over time consistently.
Also, it should be discriminative enough to prevent it from being matched to other object instances.
This can be achieved with the contrastive learning objective~\cite{hadsell2006dimensionality}, where the positive can be oneself, and set the negatives other object embeddings.
However, given that the number of objects appearing in a video is limited in typical, we significantly lack training samples. As a remedy, we introduce three strategies in the following for scaling the training data.

First, we allow using positive proposals~\footnote{The proposals that have IoU greater than the given threshold with the ground truths.}, and concatenate them over the time axis to generate object tracks.
This strategy has two clear advantages: 1) we can significantly increase the number of object tracks compared to only using ground truth boxes. 2) we can enjoy the natural jittering effect of the proposals, allowing the model to be robust to temporal appearance changes(see~\figref{fig:clip_tracker}-(1)).

Second, we simulate hard positives and negatives.
The first is mislocalization due to some boxes along the track that does not tightly cover the object.
The second is a track drifting due to the incorrect match to other instances in the middle of the track.
To make the model robust to these errors, we augment the training samples with the following two new strategies.

\begin{enumerate}
\setlength\itemsep{0.5em}
\item \textbf{Negative Proposals.}
For the localization error, we incorporate negative proposals that have loose IoU with the ground truths (in between 0.3 and 0.5).
Given both the positive and negative proposals, three different types of object tracks are generated in equal probability during training; positive only, negative only, and hybrid.

\item \textbf{Object Track Mixup.}
For the track drifting error, we propose to mix two different object tracks~\cite{zhang2017mixup}.
In particular, we switch random portions of the given object track embeddings with randomly selected different object track embeddings.
To keep the original identity of the given track, we bound the mixup ratio to 0.3.
\end{enumerate}

With these two new augmentation strategies, we can effectively simulate hard positives and negatives, which can help the model improve embedding quality (see~\figref{fig:clip_tracker}-(2) and (3)).
The proposal sampling is conducted in an order-free manner, which means the object tracks can be constructed with multiple boxes from the same image.

Finally, we treat the object instances from other videos in the training batch as negatives. This naturally increases the negatives in the contrastive learning context, leading to more discriminative feature embeddings~\cite{chen2020simple,he2020momentum}.

\vspace{2mm}
\noindent{\textbf{Objective Function.}}
Under this setup, we can apply the following contrastive learning formula,
\begin{equation}
    \begin{split}
    \mathcal{L} &= \mathrm{log}{[1 + \mathrm{\sum_{\mathbf{z^{+}}}\sum_{\mathbf{z^{-}}}exp(\mathbf{z}\cdot \mathbf{z^{-}} - \mathbf{z}\cdot \mathbf{z^{+}})}]}. \\
    \end{split}
\end{equation}
Where $\mathrm{\mathbf{z}}$ denotes an anchor, $\mathrm{\mathbf{z^{+}}}$ and $\mathrm{\mathbf{z^{-}}}$  are corresponding positives and negatives. We note the formula allows multiple positives, which is an extension of the standard single-positive formula, and thus can learn more discriminative embeddings~\cite{pang2021quasi}.

\vspace{2mm}
\noindent{\textbf{Track History Management.}}
At test-time, the clip tracker holds the global video-level object tracks and expands those incrementally by merging the outputs of the intra-clip tracker.
In specific, the affinity matrix is computed based on the \textit{summary} feature similarities between clip and global tracks.
We associate the tracks with the matching threshold of 0.5.
If the match is found, we save all the frame-level predictions of the objects in the memory buffer for the subsequent matching.

\vspace{2mm}
\noindent{\textbf{Memory Buffer Size.}}
In computing the summary embedding of the global tracks, we set a proper memory buffer size. Intuitively, the larger the memory buffer size, the more the model can exploit the temporal context up to its temporal abstraction capacity.
In practice, we set to 30 and 10 frames for MOT17 and TAO, respectively.

\section{Experiments}

In this section, we conduct experiments to demonstrate the benefits of our proposals using the TAO~\cite{dave2020tao} and MOT17~\cite{milan2016mot16} benchmarks.
We investigate the results focusing mainly on the association aspect.
Specifically, we use Track AP for the evaluation on the TAO benchmark. 
Also, Track AP$_{S}$, Track AP$_{M}$, and Track AP$_{L}$ are adopted which indicates the Track AP of (short $\leq 3$), ($3\leq$ medium $\leq10$), and (long $\geq 10$) length object tracks.
We use MOTA~\cite{bernardin2008evaluating}, IDF1~\cite{ristani2016performance}, and the recently presented HOTA metrics~\cite{luiten2021hota} for the evaluation on the MOT17 benchmark. To see the association quality more thoroughly, we also present $\mathrm{AssA}$ performance.

\subsection{Datasets}

\vspace{2mm}
\noindent{\textbf{TAO}~\cite{dave2020tao}} 
is the first video benchmark for large vocabulary object tracking.
It annotates 482 classes in total, which are the subset of the LVIS~\cite{gupta2019lvis} dataset.
Its training set has 400 videos, covering 216 classes. The validation set consists of 988 videos, spanning 302 classes. The test set has 1419 videos, covering 369 classes.
The videos are annotated in 1 FPS.

\vspace{2mm}
\noindent{\textbf{MOT17}~\cite{milan2016mot16}} 
is the most popular multi-object tracking benchmark, which consists of 7 training and 7 testing videos.
While it has only pedestrian annotations, sequences include several challenges such as frequent occlusions and crowd scenes.
The video framerate is relatively high, ranging from 14 to 30 FPS.
For ablation experiments, we split each training sequence into two halves, and use the first half-frames for training and the second for validation.

\vspace{2mm}
\noindent Due to the difference in frame rate (1FPS \textit{v.s.} 30FPS) and annotated object categories (482 \textit{v.s.} 1), TAO and MOT17 cover significantly different motion patterns and video contents.
We show that our proposal works well on both.

\subsection{Implementation Details}

\vspace{2mm}
\noindent{\textbf{Clip Tracker Architecture.}} \hspace{1mm}
The Clip Tracker model consists of 3 transformer encoder layers, where each layer has 8 heads with embedding dimensions of 256 (for MOT17) and 512 (for TAO).
The input of Clip Tracker is the track embedding sequence of each object instance and the trainable track token.
We apply a single linear layer to embed the input sequence. 
Finally, the output of the track token passes through a single MLP layer to produce the summary of the object track history.
The positional encodings are not used in this work.

\vspace{2mm}
\noindent{\textbf{Hyperparameters.}} \hspace{1mm}
The proposed clip-based tracking pipeline introduces two new hyperparameters, clip size and clip interval (see~\algref{alg:clip_track}).
The clip size and interval directly affect the video chunking patterns.
We observe that overlapping the clip prediction eases the inter-clip association. 
In practice, we considered the frame rate of the given datasets.
For example, for the TAO, which provides low frame rate video inputs, we set clip size and interval value to 6 and 3, respectively.
For the MOT, we set clip size and interval to 10 and 5, respectively.

\vspace{2mm}
\noindent{\textbf{Training.}} \hspace{1mm}
We use MMdetection and MMTracking frameworks~\cite{chen2019mmdetection}.
For the TAO experiments, COCO-style training schedule of 2$\times$ and 1$\times$ are adopted for LVIS pre-training and TAO fine-tuning, respectively. 
For the MOT experiments, we pre-trained the model using CrowdHuman dataset~\cite{shao2018crowdhuman} for 12 epochs and then fine-tuned on the MOT17 train for 4 epochs.
During pre-training, the detection part is learned.
A batch size of 16 (2 per GPU) and an initial learning rate of 0.02 are used.
We use the backbone ResNet-101 (ResNet-50) for the TAO (MOT17) experiment following the previous studies~\cite{dave2020tao,pang2021quasi}.
We train the Clip Tracker by sampling 2 and 6 frames under the frame range of 1 and 30 for TAO and MOT17 experiments, respectively.

\vspace{2mm}
\noindent{\textbf{Testing.}} \hspace{1mm}
Our method processes video frames recursively at clip-level, generating object tracks for the given clip and merging them into the global tracks incrementally. 
We use standard directional matching for intra- and Clip-Tracker for inter-clip tracking.
The overall tracking procedure follows~\algref{alg:clip_track}.
While the frame-level association capability within the clip is similar to that of the baseline, video chunking along with the clip-wise association allows tracking performance improvement.
We use resized frames of 1080$\times$1080 for testing.

\begin{table*}[t!]
    \centering
\setlength{\tabcolsep}{5pt}
{\renewcommand\arraystretch{0.95}
    \resizebox{\textwidth}{!}
    {
        \begin{tabular}{l|ccc|ccc}
        \toprule
        \multicolumn{1}{c|}{} & \multicolumn{3}{c|}{\textbf{TAO val}} & \multicolumn{3}{c}{\textbf{TAO test}}\\ 
        \multicolumn{1}{c|}{Method} & $\mathrm{TrackAP}_{50}$ & $\mathrm{TrackAP}_{75}$ & $\mathrm{TrackAP}_{50:95}$ & $\mathrm{TrackAP}_{50}$ & $\mathrm{TrackAP}_{75}$ & $\mathrm{TrackAP}_{50:95}$ \\
        \toprule
            SORT~\cite{dave2020tao}       &13.2 & -   & -     & 10.2 & 4.4 & 4.9 \\
            QDTrack~\cite{pang2021quasi}  &13.4 & 4.9 & 6.1   & 12.6 & 4.5 & 5.6 \\
            \midrule 
            Ours    &\textbf{17.7} & \textbf{5.8} & \textbf{7.3}   & \textbf{15.8} & \textbf{5.9} & \textbf{6.6}  \\ 
        \bottomrule
        \end{tabular}
    }
}
\vspace{2mm}
\captionsetup{font=footnotesize}
\caption{Tracking Results on TAO \textit{val} and \textit{test}. We use clip size and interval of 6 and 3, respectively.}
\label{tab:tao_sota}
\vspace{-6mm}
\end{table*}
\begin{table*}[t!]
\centering
\setlength{\tabcolsep}{15pt}
\centering
{\renewcommand\arraystretch{1}
    \resizebox{\textwidth}{!}
    {
        \begin{tabular}{l|ccc|ccc}
        \toprule
        \multicolumn{1}{c|}{} & \multicolumn{3}{c|}{\textbf{MOT val}} & \multicolumn{3}{c}{\textbf{MOT test}}\\ 
        \multicolumn{1}{c|}{Method} & MOTA & IDF1 & HOTA & MOTA & IDF1 & HOTA \\
        \toprule
            QDTrack~\cite{pang2021quasi}  &68.1 & 69.5 & 57.5   & 71.1 & 70.2 & 57.8 \\
            \midrule 
            Ours    & \textbf{68.9} & \textbf{72.4} & \textbf{59.7}   & \textbf{71.6} & \textbf{72.7} & \textbf{59.0}  \\ 
        \bottomrule
        \end{tabular}
    }
}
    \vspace{2mm}
    \captionsetup{font=footnotesize}
    \caption{Tracking Results on MOT17. We use clip size and interval of 10 and 5, respectively.}
    \label{tab:mot_sota}
    \vspace{-3mm}
\hfill
\end{table*}

\subsection{Main Results}
Upon the state-of-the-art tracking-by-detection framework, QDTrack~\cite{pang2021quasi}, we instantiate our proposal, clip-based tracking.
The main results on TAO and MOT17 are summarized in~\tabref{tab:tao_sota} and~\tabref{tab:mot_sota}.
We observe our proposal pushes the tracking performances on both datasets.
The results imply that the clip-based association is generic over various scenarios.

\subsection{Ablation Studies}
To empirically confirm the effectiveness of our proposals, we conduct  ablation studies using the TAO and MOT17 benchmarks.
We focus on the association quality; thus, we use the metrics of IDF1, HOTA, and ASSA for MOT, and Track AP for TAO.

\vspace{3mm}
\noindent{\textbf{Simple Clip-based Tracker.}} \hspace{1mm}
We first analyze how the standard QDTrack framework~\cite{pang2021quasi} improves by changing the tracking scheme from frame-based to clip-based.
To do so, we design a simple clip-based tracking baseline, which takes the clip as input, performs intra-clip tracking, and links the clip predictions based on the temporal averaged feature matching.
The results are shown in~\tabref{tab:intra_abl}.
We observe that the clip-based tracking scheme improves the association quality in the MOT17 benchmark but not TAO.
We see this because the motion change is significant in TAO due to the low frame rate; thus, the simple temporal averaged feature performs poorly, which motivates the design of Clip Tracker.

\vspace{3mm}
\noindent{\textbf{Intra-clip association.}} \hspace{1mm}
Among the intra-clip variants, we observe that the standard scheme is already competitive and runs in a much faster time.
Here, we use directional matching with memory mechanism for the intra-clip tracking.

\vspace{3mm}
\noindent{\textbf{Inter-clip association.}} \hspace{1mm}
We perform ablations on inter-clip association methods. 
The results are summarized in~\tabref{tab:inter_abl}.
We observe that IoU-based chaining performs inferior, even lower than the baseline.
This is because the method misses all the object tracks not in the overlapping frame, leading to significant track fragmentation.
Finally, the presented Clip Tracker significantly outperforms the simple temporal averaged feature matching.
The results show that learning-based temporal pooling is more robust than simple averaging.
We use Clip Tracker as our inter-clip association method in the following experiments.

\begin{table*}[t!]
 \centering
 \subfloat[\scriptsize Intra Clip Association Methods]{
     \label{tab:intra_abl}
     \resizebox{0.48\textwidth}{!}{
     \setlength{\tabcolsep}{3pt}
     \def\arraystretch{1.6}
     \begin{tabular}{l|  c c c | c c c c}
        \toprule
        \multicolumn{1}{c|}{}                      &\multicolumn{3}{c|}{MOT17} & \multicolumn{4}{c}{TAO}        \\
         Method                                   & IDF1 & HOTA     & AssA   & $\mathrm{T}_{50}$ & $\mathrm{T}_{S}$ & $\mathrm{T}_{M}$ & $\mathrm{T}_{L}$ \\
        \midrule
        QDtrack~\cite{pang2021quasi}              & 69.5    & 57.5   & 58.0  & 13.4  & 6.7  & 9.6     & 17.2  \\ 
        \midrule
        Direc-free~\cite{zhang2017multi}          & 70.3    & 57.9   & 58.9  & 13.1  & 6.1  & 8.9     & 16.8  \\
        Direc                                     & 70.3    & 58.6   & 59.4  & 12.3   & 5.1 & 8.3     & 16.5  \\
        \bottomrule
        \end{tabular}
     }
 }
 \subfloat[\scriptsize Inter Clip Association Methods]{
     \label{tab:inter_abl}
     \resizebox{0.48\textwidth}{!}{
     \setlength{\tabcolsep}{3.8pt}
     \def\arraystretch{1.45}
     \begin{tabular}{l | c c c | c c c c}
        \toprule
        \multicolumn{1}{c|}{}                      &\multicolumn{3}{c|}{MOT17} & \multicolumn{4}{c}{TAO}       \\
        Method                                    & IDF1 & HOTA     & AssA  & $\mathrm{T}_{50}$ & $\mathrm{T}_{S}$ & $\mathrm{T}_{M}$ & $\mathrm{T}_{L}$ \\
        \midrule
        QDtrack~\cite{pang2021quasi}              & 69.5    & 57.5   & 58.0 & 13.4  & 6.7   & 9.6  & 17.2    \\ 
        \midrule
        IoU-chain~\cite{peng2020chained}          & 69.1    & 56.4   & 56.1 & 10.2  & 3.4   & 6.5   & 14.4   \\ 
        Tmp avg.                                  & 70.3    & 58.6   & 59.4 & 12.3  & 5.1   & 8.3   & 16.8   \\ 
        Clip-Tracker                              & 72.4    & 59.7   & 61.9 & 17.7  & 10.9  & 14.6  & 21.3   \\ 
        \bottomrule
        \end{tabular}
     }
}
\vspace{2mm}
\captionsetup{font=footnotesize}
\caption{
\textbf{Impact of Clip-based Tracking.} 
We use both TAO \textit{val} and MOT \textit{val} to evaluate the impact of the new clip-based tracking approach.
We investigate various feasible implementations for both intra and inter clip associations described in~\secref{sec:intra} and~\secref{sec:inter}.
Here, T is an abbreviation of Track AP.
}
\label{tab:intra_inter_abl}
\vspace{-6mm}
\end{table*}

\begin{table*}[t!]
 \centering
\subfloat[\scriptsize Clip Size and Interval]{
     \label{tab:hyper}
     \resizebox{0.33\textwidth}{!}{
     \setlength{\tabcolsep}{2pt}
     \def\arraystretch{1.5}
     \begin{tabular}{l | l c c c }
        \toprule
        Method                                    & $[\mathrm{C}_{S}, \mathrm{C}_{I}]$               & IDF1 & HOTA & AssA        \\
        \midrule
        QDtrack~\cite{pang2021quasi}              &                 & 69.5    & 57.5 & 58.0     \\ 
        \midrule
        \multirow{4}{*}{Clip-Tracker}             & [5,5]           & 71.6    & 59.0 & 60.5     \\
                                                  & [10,10]         & 71.1    & 58.6 & 59.8     \\
                                                  & [10,5]          & 72.4    & 59.7 & 61.9     \\
                                                  & [20,10]         & 71.9    & 59.5 & 61.3     \\
        \bottomrule
        \end{tabular}
     }
}
 \subfloat[\scriptsize Track Augmentations]{
     \label{tab:track_aug}\hspace{-3mm}
     \resizebox{0.33\textwidth}{!}{
     \setlength{\tabcolsep}{3pt}
     \def\arraystretch{2.35}
     \begin{tabular}{c c | c c c }
        \toprule
        Loc. Err & Drift. Err                     & IDF1 & HOTA     & AssA   \\
        \midrule
                                &                 & 70.3 & 58.8     & 60.2    \\
        \checkmark &                              & 71.9 & 59.3     & 61.3   \\
        \checkmark & \checkmark                   & 72.4 & 59.7     & 61.9   \\
        \bottomrule
        \end{tabular}
     }
 }
 \subfloat[\scriptsize Global Track Management]{
     \label{tab:track_hist}\hspace{-3mm}
     \resizebox{0.33\textwidth}{!}{
     \setlength{\tabcolsep}{2pt}
     \def\arraystretch{2.35}
     \begin{tabular}{l | c c c }
        \toprule
        Method                                    & IDF1 & HOTA     & AssA  \\
        \midrule
        Two-clip based.                          & 71.8 & 59.2     & 60.4  \\
        Moving avg.                               & 71.7 & 59.3     & 60.9  \\
        Feature bank. (Ours)                       & 72.4 & 59.7     & 61.9  \\
        \bottomrule
        \end{tabular}
     }
}
\vspace{3mm}
\captionsetup{font=footnotesize}
\caption{
\textbf{Component Analysis on Clip Tracker} using MOT \textit{val}.
First, we conduct parameter analysis on two newly introduced hyperparameters: clip size $\mathrm{C}_{S}$ and clip interval $\mathrm{C}_{I}$.
Second, We study the impact of two track augmentation strategies, negative proposal sampling (for localization error simulation) and object track mixup (for track drifting simulation), during the Clip Tracker training.
Lastly, we investigate three different baseline methods to manage global video-level tracks.
}
\label{tab:comp anal}
\vspace{-3mm}
\end{table*}

\begin{figure}[t!]
    \centering 
    \includegraphics[width=0.98\textwidth]{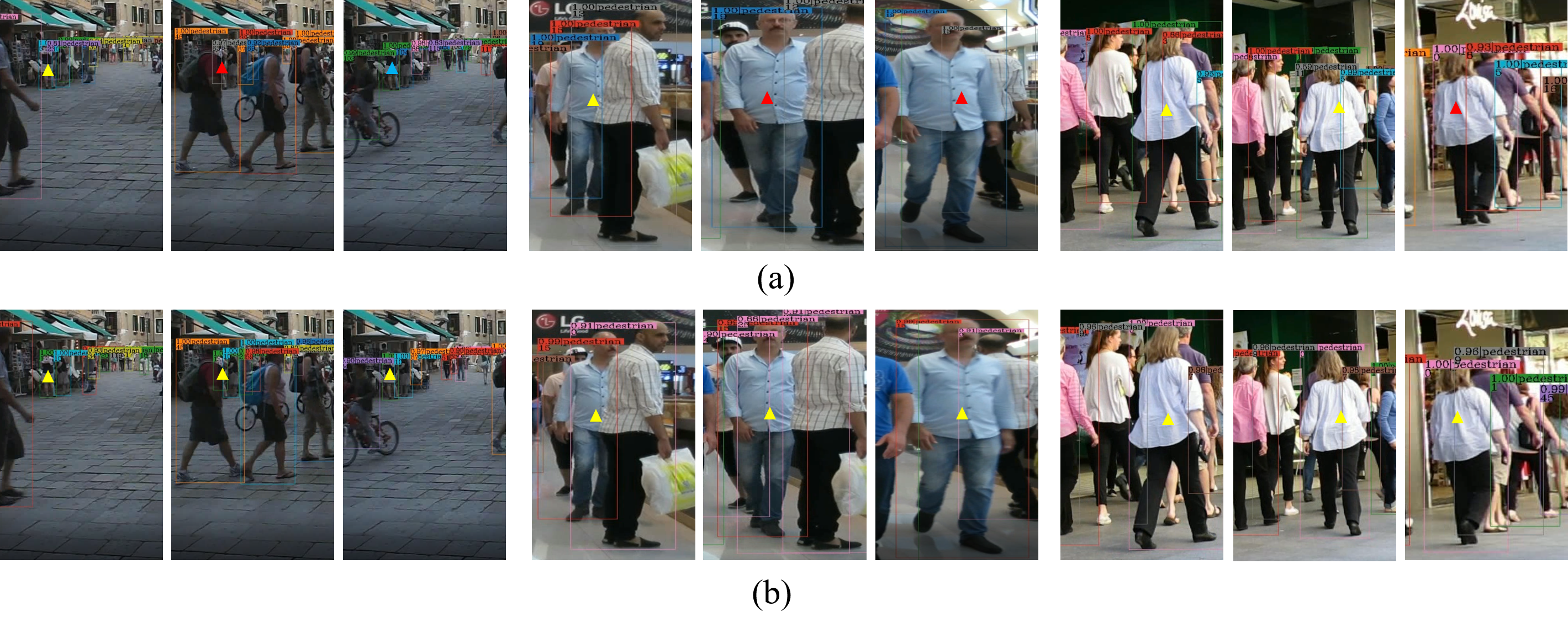}
    \vspace{-3mm}
    \captionsetup{font=footnotesize}
    \caption{
    \textbf{Qualitative Results.}
    (a) Tracking by detection. (b) Tracking by associating clips. 
    The colored triangle represents the predicted person id.
    \textit{Best Viewed in Colors.}
    }
    \vspace{-3mm}
    \label{fig:qual}
\end{figure}

\vspace{3mm}
\noindent{\textbf{Parameter Analysis on Clip Size and Interval.}} \hspace{1mm}
We conducted parameter analysis on the two newly introduced hyperparameters, clip size and interval.
These hyperparameters affect the video chunking pattern and, as a result, impact both the intra- and inter-clip association quality.
The results are summarized in~\tabref{tab:hyper}.
We observe two tendencies.
First, the association improves when the overlapping exists in-between the clips, \textit{i.e.} $\mathrm{C}_{S} > \mathrm{C}_{I}$.
Intuitively, the overlapping makes the feature embedding sequence quite similar, and thus the Clip Tracker can more easily find the correct match.
Second, we see a slight performance drop when the clip size increases, \textit{e.g.} $[5,5] \rightarrow [10,10]$ or $[10,5] \rightarrow [20,10]$.
We see this phenomenon is mainly due to the increased false positives with the large clip size, and we mitigate this issue by introducing the overlapping frames, $[10,10] \rightarrow [10,5]$.
In practice, proper clip size and interval values should be set, and here we found the clip size of 10 and the interval of 5 strikes a balance.

\vspace{3mm}
\noindent{\textbf{Track Augmentations.}} \hspace{1mm}
In order to improve the robustness of the Clip Tracker, we introduced two new hard training sample generation strategies, negative proposal sampling and object track mixup.
We investigate their impacts on the tracking performance in~\tabref{tab:track_aug}.
From the baseline of using only the positive proposals, the track performance improves meaningfully with the negative proposal sampling.
The object track mixup further improves the score.

\vspace{3mm}
\noindent{\textbf{Track Management Strategies.}} \hspace{1mm}
The inter-clip tracker manages global video-level track over time (see~\algref{alg:clip_track}).
It saves the tracked object's history for matching and merging subsequent clip prediction.
Here, we examine three different feasible implementations for this.
First is a two-clip-based matching, which only saves the immediately preceding object track features.
The second is a moving-average-based matching, which uses smoothed object prototype embeddings.
The third is feature bank based matching, which saves all the object track features up to pre-setted buffer size.
As can be shown in~\tabref{tab:track_hist}, we observe that the feature bank approach produces the best performance.
This is because it has the broadest temporal view and can directly utilize the raw feature sequence rather than the temporally diluted features. 

\subsection{Qualitative Results}
The visual comparisons of tracking results between the standard tracking scheme and our proposal are in~\figref{fig:qual}.
We select 3 different difficult sequences from the half validation set of MOT17.
This include small objects, occlusion, and crowd scene.
We see that our approach consistently tracks the objects without identity changes for long-term.

\section{Conclusion}
In this paper, we point out the fundamental limitations of the current tracking-by-detection scheme.
Due to its sequential frame-based matching property, the method suffers from intermediate disturbances in a video and overlooks the temporal context around the target matching frame.
As an alternative, we present a new tracking method based on the clip.
For the implementation, we define two new tracking operations, intra- and inter-clip tracking.
The new formulation views a single long video sequence as multiple short clips.
In this way, the intra-clip tracking turns out to be short-term tracking, and we observe that the standard tracking approaches with memory are already competitive in this setup. 
For the inter-clip tracking, feature-based matching is more stable than IoU-based chaining.
We further improve the association quality by designing a novel transformer-based approach, namely Clip Tracker.
By connecting together, we show that our proposal achieves new state-of-the-art results on the challenging large vocabulary object tracking benchmark, TAO.
We also confirm that our method significantly improves the association quality of the baseline in MOT17.
We hope our work opens a new view of the current tracking paradigm.

\vspace{3mm}
\noindent\textbf{Acknowledgement}
This work was supported in part by the National Research Foundation of Korea (NRF-2020M3H8A1115028, FY2021).

\clearpage
%
%
\bibliographystyle{splncs04}
\bibliography{egbib}

\begin{thebibliography}{10}
\providecommand{\url}[1]{\texttt{#1}}
\providecommand{\urlprefix}{URL }
\providecommand{\doi}[1]{https://doi.org/#1}

\bibitem{andriyenko2011multi}
Andriyenko, A., Schindler, K.: Multi-target tracking by continuous energy
  minimization. In: CVPR 2011. pp. 1265--1272. IEEE (2011)

\bibitem{andriyenko2012discrete}
Andriyenko, A., Schindler, K., Roth, S.: Discrete-continuous optimization for
  multi-target tracking. In: 2012 IEEE Conference on Computer Vision and
  Pattern Recognition. pp. 1926--1933. IEEE (2012)

\bibitem{athar2020stem}
Athar, A., Mahadevan, S., Osep, A., Leal-Taix{\'e}, L., Leibe, B.: Stem-seg:
  Spatio-temporal embeddings for instance segmentation in videos. In: European
  Conference on Computer Vision. pp. 158--177. Springer (2020)

\bibitem{bergmann2019tracking}
Bergmann, P., Meinhardt, T., Leal-Taixe, L.: Tracking without bells and
  whistles. In: Proceedings of the IEEE/CVF International Conference on
  Computer Vision. pp. 941--951 (2019)

\bibitem{bernardin2008evaluating}
Bernardin, K., Stiefelhagen, R.: Evaluating multiple object tracking
  performance: the clear mot metrics. EURASIP Journal on Image and Video
  Processing  \textbf{2008},  1--10 (2008)

\bibitem{bewley2016simple}
Bewley, A., Ge, Z., Ott, L., Ramos, F., Upcroft, B.: Simple online and realtime
  tracking. In: 2016 IEEE international conference on image processing (ICIP).
  pp. 3464--3468. IEEE (2016)

\bibitem{carion2020end}
Carion, N., Massa, F., Synnaeve, G., Usunier, N., Kirillov, A., Zagoruyko, S.:
  End-to-end object detection with transformers. In: European conference on
  computer vision. pp. 213--229. Springer (2020)

\bibitem{chen2019mmdetection}
Chen, K., Wang, J., Pang, J., Cao, Y., Xiong, Y., Li, X., Sun, S., Feng, W.,
  Liu, Z., Xu, J., et~al.: Mmdetection: Open mmlab detection toolbox and
  benchmark. arXiv preprint arXiv:1906.07155  (2019)

\bibitem{chen2018real}
Chen, L., Ai, H., Zhuang, Z., Shang, C.: Real-time multiple people tracking
  with deeply learned candidate selection and person re-identification. In:
  2018 IEEE international conference on multimedia and expo (ICME). pp.~1--6.
  IEEE (2018)

\bibitem{chen2020simple}
Chen, T., Kornblith, S., Norouzi, M., Hinton, G.: A simple framework for
  contrastive learning of visual representations. In: International conference
  on machine learning. pp. 1597--1607. PMLR (2020)

\bibitem{dave2020tao}
Dave, A., Khurana, T., Tokmakov, P., Schmid, C., Ramanan, D.: Tao: A
  large-scale benchmark for tracking any object. In: ECCV. pp. 436--454.
  Springer (2020)

\bibitem{dosovitskiy2020image}
Dosovitskiy, A., Beyer, L., Kolesnikov, A., Weissenborn, D., Zhai, X.,
  Unterthiner, T., Dehghani, M., Minderer, M., Heigold, G., Gelly, S., et~al.:
  An image is worth 16x16 words: Transformers for image recognition at scale.
  arXiv preprint arXiv:2010.11929  (2020)

\bibitem{feichtenhofer2017detect}
Feichtenhofer, C., Pinz, A., Zisserman, A.: Detect to track and track to
  detect. In: Proceedings of the IEEE international conference on computer
  vision. pp. 3038--3046 (2017)

\bibitem{fortmann1983sonar}
Fortmann, T., Bar-Shalom, Y., Scheffe, M.: Sonar tracking of multiple targets
  using joint probabilistic data association. IEEE journal of Oceanic
  Engineering  \textbf{8}(3),  173--184 (1983)

\bibitem{gupta2019lvis}
Gupta, A., Dollar, P., Girshick, R.: Lvis: A dataset for large vocabulary
  instance segmentation. In: CVPR. pp. 5356--5364 (2019)

\bibitem{hadsell2006dimensionality}
Hadsell, R., Chopra, S., LeCun, Y.: Dimensionality reduction by learning an
  invariant mapping. In: 2006 IEEE Computer Society Conference on Computer
  Vision and Pattern Recognition (CVPR'06). vol.~2, pp. 1735--1742. IEEE (2006)

\bibitem{he2020momentum}
He, K., Fan, H., Wu, Y., Xie, S., Girshick, R.: Momentum contrast for
  unsupervised visual representation learning. In: Proceedings of the IEEE/CVF
  conference on computer vision and pattern recognition. pp. 9729--9738 (2020)

\bibitem{hu2019joint}
Hu, H.N., Cai, Q.Z., Wang, D., Lin, J., Sun, M., Krahenbuhl, P., Darrell, T.,
  Yu, F.: Joint monocular 3d vehicle detection and tracking. In: Proceedings of
  the IEEE/CVF International Conference on Computer Vision. pp. 5390--5399
  (2019)

\bibitem{hwang2021video}
Hwang, S., Heo, M., Oh, S.W., Kim, S.J.: Video instance segmentation using
  inter-frame communication transformers. Advances in Neural Information
  Processing Systems  \textbf{34} (2021)

\bibitem{janai2018unsupervised}
Janai, J., Guney, F., Ranjan, A., Black, M., Geiger, A.: Unsupervised learning
  of multi-frame optical flow with occlusions. In: Proceedings of the European
  Conference on Computer Vision (ECCV). pp. 690--706 (2018)

\bibitem{leal2016learning}
Leal-Taix{\'e}, L., Canton-Ferrer, C., Schindler, K.: Learning by tracking:
  Siamese cnn for robust target association. In: Proceedings of the IEEE
  Conference on Computer Vision and Pattern Recognition Workshops. pp. 33--40
  (2016)

\bibitem{leal2017tracking}
Leal-Taix{\'e}, L., Milan, A., Schindler, K., Cremers, D., Reid, I., Roth, S.:
  Tracking the trackers: an analysis of the state of the art in multiple object
  tracking. arXiv:1704.02781  (2017)

\bibitem{liang2020rethinking}
Liang, C., Zhang, Z., Lu, Y., Zhou, X., Li, B., Ye, X., Zou, J.: Rethinking the
  competition between detection and reid in multi-object tracking. arXiv
  preprint arXiv:2010.12138  (2020)

\bibitem{liu2021swin}
Liu, Z., Lin, Y., Cao, Y., Hu, H., Wei, Y., Zhang, Z., Lin, S., Guo, B.: Swin
  transformer: Hierarchical vision transformer using shifted windows. In:
  Proceedings of the IEEE/CVF International Conference on Computer Vision. pp.
  10012--10022 (2021)

\bibitem{luiten2021hota}
Luiten, J., Osep, A., Dendorfer, P., Torr, P., Geiger, A., Leal-Taix{\'e}, L.,
  Leibe, B.: Hota: A higher order metric for evaluating multi-object tracking.
  International journal of computer vision  \textbf{129}(2),  548--578 (2021)

\bibitem{meinhardt2021trackformer}
Meinhardt, T., Kirillov, A., Leal-Taixe, L., Feichtenhofer, C.: Trackformer:
  Multi-object tracking with transformers. arXiv preprint arXiv:2101.02702
  (2021)

\bibitem{milan2016mot16}
Milan, A., Leal-Taix{\'e}, L., Reid, I., Roth, S., Schindler, K.: Mot16: A
  benchmark for multi-object tracking. arXiv preprint arXiv:1603.00831  (2016)

\bibitem{pang2021quasi}
Pang, J., Qiu, L., Li, X., Chen, H., Li, Q., Darrell, T., Yu, F.: Quasi-dense
  similarity learning for multiple object tracking. In: Proceedings of the
  IEEE/CVF conference on computer vision and pattern recognition. pp. 164--173
  (2021)

\bibitem{park2022per}
Park, K., Woo, S., Oh, S.W., Kweon, I.S., Lee, J.Y.: Per-clip video object
  segmentation. In: Proceedings of the IEEE/CVF Conference on Computer Vision
  and Pattern Recognition. pp. 1352--1361 (2022)

\bibitem{parmar2018image}
Parmar, N., Vaswani, A., Uszkoreit, J., Kaiser, L., Shazeer, N., Ku, A., Tran,
  D.: Image transformer. In: International Conference on Machine Learning. pp.
  4055--4064. PMLR (2018)

\bibitem{peng2020chained}
Peng, J., Wang, C., Wan, F., Wu, Y., Wang, Y., Tai, Y., Wang, C., Li, J.,
  Huang, F., Fu, Y.: Chained-tracker: Chaining paired attentive regression
  results for end-to-end joint multiple-object detection and tracking. In:
  European conference on computer vision. pp. 145--161. Springer (2020)

\bibitem{ramanan2003finding}
Ramanan, D., Forsyth, D.A.: Finding and tracking people from the bottom up. In:
  CVPR. vol.~2, pp. II--II. IEEE (2003)

\bibitem{redmon2016you}
Redmon, J., Divvala, S., Girshick, R., Farhadi, A.: You only look once:
  Unified, real-time object detection. In: Proceedings of the IEEE conference
  on computer vision and pattern recognition. pp. 779--788 (2016)

\bibitem{ren2015faster}
Ren, S., He, K., Girshick, R., Sun, J.: Faster r-cnn: Towards real-time object
  detection with region proposal networks. Advances in neural information
  processing systems  \textbf{28} (2015)

\bibitem{rezatofighi2015joint}
Rezatofighi, S.H., Milan, A., Zhang, Z., Shi, Q., Dick, A., Reid, I.: Joint
  probabilistic data association revisited. In: Proceedings of the IEEE
  international conference on computer vision. pp. 3047--3055 (2015)

\bibitem{ristani2016performance}
Ristani, E., Solera, F., Zou, R., Cucchiara, R., Tomasi, C.: Performance
  measures and a data set for multi-target, multi-camera tracking. In: European
  conference on computer vision. pp. 17--35. Springer (2016)

\bibitem{sadeghian2017tracking}
Sadeghian, A., Alahi, A., Savarese, S.: Tracking the untrackable: Learning to
  track multiple cues with long-term dependencies. In: Proceedings of the IEEE
  international conference on computer vision. pp. 300--311 (2017)

\bibitem{shao2018crowdhuman}
Shao, S., Zhao, Z., Li, B., Xiao, T., Yu, G., Zhang, X., Sun, J.: Crowdhuman: A
  benchmark for detecting human in a crowd. arXiv preprint arXiv:1805.00123
  (2018)

\bibitem{streit1994maximum}
Streit, R.L., Luginbuhl, T.E.: Maximum likelihood method for probabilistic
  multihypothesis tracking. In: Signal and Data Processing of Small Targets
  1994. vol.~2235, pp. 394--405. International Society for Optics and Photonics
  (1994)

\bibitem{sun2020transtrack}
Sun, P., Cao, J., Jiang, Y., Zhang, R., Xie, E., Yuan, Z., Wang, C., Luo, P.:
  Transtrack: Multiple object tracking with transformer. arXiv preprint
  arXiv:2012.15460  (2020)

\bibitem{sun2019deep}
Sun, S., Akhtar, N., Song, H., Mian, A., Shah, M.: Deep affinity network for
  multiple object tracking. IEEE transactions on pattern analysis and machine
  intelligence  \textbf{43}(1),  104--119 (2019)

\bibitem{vaswani2017attention}
Vaswani, A., Shazeer, N., Parmar, N., Uszkoreit, J., Jones, L., Gomez, A.N.,
  Kaiser, {\L}., Polosukhin, I.: Attention is all you need. Advances in neural
  information processing systems  \textbf{30} (2017)

\bibitem{wang2021end}
Wang, Y., Xu, Z., Wang, X., Shen, C., Cheng, B., Shen, H., Xia, H.: End-to-end
  video instance segmentation with transformers. In: Proceedings of the
  IEEE/CVF Conference on Computer Vision and Pattern Recognition. pp.
  8741--8750 (2021)

\bibitem{wang2020towards}
Wang, Z., Zheng, L., Liu, Y., Li, Y., Wang, S.: Towards real-time multi-object
  tracking. In: European Conference on Computer Vision. pp. 107--122. Springer
  (2020)

\bibitem{watson2021temporal}
Watson, J., Mac~Aodha, O., Prisacariu, V., Brostow, G., Firman, M.: The
  temporal opportunist: Self-supervised multi-frame monocular depth. In:
  Proceedings of the IEEE/CVF Conference on Computer Vision and Pattern
  Recognition. pp. 1164--1174 (2021)

\bibitem{wojke2017simple}
Wojke, N., Bewley, A., Paulus, D.: Simple online and realtime tracking with a
  deep association metric. In: 2017 IEEE international conference on image
  processing (ICIP). pp. 3645--3649. IEEE (2017)

\bibitem{wu2021track}
Wu, J., Cao, J., Song, L., Wang, Y., Yang, M., Yuan, J.: Track to detect and
  segment: An online multi-object tracker. In: Proceedings of the IEEE/CVF
  conference on computer vision and pattern recognition. pp. 12352--12361
  (2021)

\bibitem{wu2021seqformer}
Wu, J., Jiang, Y., Zhang, W., Bai, X., Bai, S.: Seqformer: a frustratingly
  simple model for video instance segmentation. arXiv preprint arXiv:2112.08275
   (2021)

\bibitem{zeng2021motr}
Zeng, F., Dong, B., Wang, T., Zhang, X., Wei, Y.: Motr: End-to-end
  multiple-object tracking with transformer. arXiv preprint arXiv:2105.03247
  (2021)

\bibitem{zhang2017mixup}
Zhang, H., Cisse, M., Dauphin, Y.N., Lopez-Paz, D.: mixup: Beyond empirical
  risk minimization. arXiv preprint arXiv:1710.09412  (2017)

\bibitem{zhang2021bytetrack}
Zhang, Y., Sun, P., Jiang, Y., Yu, D., Yuan, Z., Luo, P., Liu, W., Wang, X.:
  Bytetrack: Multi-object tracking by associating every detection box. arXiv
  preprint arXiv:2110.06864  (2021)

\bibitem{zhang2021fairmot}
Zhang, Y., Wang, C., Wang, X., Zeng, W., Liu, W.: Fairmot: On the fairness of
  detection and re-identification in multiple object tracking. International
  Journal of Computer Vision  \textbf{129}(11),  3069--3087 (2021)

\bibitem{zhang2017multi}
Zhang, Z., Wu, J., Zhang, X., Zhang, C.: Multi-target, multi-camera tracking by
  hierarchical clustering: Recent progress on dukemtmc project. arXiv preprint
  arXiv:1712.09531  (2017)

\bibitem{zheng2021rethinking}
Zheng, S., Lu, J., Zhao, H., Zhu, X., Luo, Z., Wang, Y., Fu, Y., Feng, J.,
  Xiang, T., Torr, P.H., et~al.: Rethinking semantic segmentation from a
  sequence-to-sequence perspective with transformers. In: Proceedings of the
  IEEE/CVF conference on computer vision and pattern recognition. pp.
  6881--6890 (2021)

\bibitem{zhou2020tracking}
Zhou, X., Koltun, V., Kr{\"a}henb{\"u}hl, P.: Tracking objects as points. In:
  European Conference on Computer Vision. pp. 474--490. Springer (2020)

\bibitem{zhu2020deformable}
Zhu, X., Su, W., Lu, L., Li, B., Wang, X., Dai, J.: Deformable detr: Deformable
  transformers for end-to-end object detection. arXiv preprint arXiv:2010.04159
   (2020)

\end{thebibliography}
\end{document}